\documentclass[sigconf,review=false,screen,authorversion,nonacm,balance=false]{acmart}

\usepackage{popets}
\usepackage{float}
\usepackage{algorithm}
\usepackage{algpseudocode}
\usepackage{subcaption}
\usepackage{caption}
\usepackage{hyperref}
\usepackage{hyperxmp}

\acmYear{YYYY}
\acmVolume{YYYY}
\acmNumber{X}
\acmDOI{XXXXXXX.XXXXXXX}
\acmISBN{}
\begin{document}

\title[Differentially Private Non Parametric Copulas]{Differentially Private Non Parametric Copulas: Generating synthetic data with non parametric copulas under privacy guarantees}


\author{Pablo A. Osorio-Marulanda}
\affiliation{%
  \institution{School of Applied Sciences and Engineering, Universidad EAFIT}
  \city{Medellín}
  \country{Colombia}}
\email{paosoriom@eafit.edu.co}

\author{John Esteban Castro Ramirez}
\affiliation{%
  \institution{School of Applied Sciences and Engineering, Universidad EAFIT}
  \city{Medellín}
  \country{Colombia}}
\email{jecastror@eafit.edu.co}

\author{Mikel Hernández Jiménez}
\affiliation{%
  \institution{Digital Health and Biomedical Technologies, Vicomtech Foundation, Basque Research and Technology Alliance (BRTA)}
  \city{Donostia - San Sebastian}
  \country{Spain}}
\email{mhernandez@vicomtech.org}

\author{Nicolas Moreno Reyes}
\affiliation{%
 \institution{School of Applied Sciences and Engineering, Universidad EAFIT}
  \city{Medellín}
  \country{Colombia}}
\email{namorenor@eafit.edu.co}

\author{Gorka Epelde Unanue}
\affiliation{%
  \institution{Digital Health and Biomedical Technologies, Vicomtech Foundation, Basque Research and Technology Alliance (BRTA)}
  \city{Donostia - San Sebastian}
  \country{Spain}
  \institution{eHealth Group, Biogipuzkoa Health Research Institute}
  \city{Donostia - San Sebastian}
  \country{Spain}
  }
\email{gepelde@vicomtech.org}


\renewcommand{\shortauthors}{Osorio-Marulanda et al.}

\begin{abstract}
  Creation of synthetic data models has represented a significant advancement across diverse scientific fields, but this technology also brings important privacy considerations for users. This work focuses on enhancing a non-parametric copula-based synthetic data generation model, DPNPC, by incorporating Differential Privacy through an Enhanced Fourier Perturbation method. The model generates synthetic data for mixed tabular databases while preserving privacy. We compare DPNPC with three other models (PrivBayes, DP-Copula, and DP-Histogram) across three public datasets, evaluating privacy, utility, and execution time. DPNPC outperforms others in modeling multivariate dependencies, maintaining privacy for small $\epsilon$ values, and reducing training times. However, limitations include the need to assess the model's performance with different encoding methods and consider additional privacy attacks. Future research should address these areas to enhance privacy-preserving synthetic data generation.
\end{abstract}

\keywords{Synthetic Data Generation, Differential Privacy, Non Parametric Copulas}

\maketitle

\section{Introduction}

The rapid growth of the technology industry, driven by the advent of the new digital revolution through Big Data, has enabled data analysis to become a crucial tool for decision-making across various fields of knowledge and industry. Alongside this trend, the technical advancement in artificial intelligence have led to the creation of synthetic data - artificially generated data produced by algorithms. This synthetic data has garnered significant interest not only in research fields but also in sectors such as medicine and health \cite{hernandez2022synthetic}, demography \cite{wang2023synthetic}, mobility \cite{berke2022generating}, education \cite{vie2022privacy}, and energy \cite{reddy1994using}.

Among the various applications of synthetic data, it is notably used to augment databases for training various machine learning models (e.g., large language models), enhance the generalization capabilities of different models \cite{beery2020synthetic,oren2021finding}, balance class distributions to ensure fairer evaluations \cite{ganganwar2024employing,treder2023approach} and anonymize information to protect privacy in the context of data sharing \cite{stadler2020synthetic}.

Synthetic data generation has gained relevance through the new Privacy Preserving Data Publishing (PPDP) frameworks, which provide methods and tools to publish useful information while preserving privacy \cite{osorio2024privacy}. However, several studies have shown that using synthetic data generation models alone is insufficient for anonymizing data in such contexts. These models are vulnerable to attacks, and artificially generated data may contain sensitive information from the original training database \cite{carlini2019secret,song2017machine}. To mitigate these threats, privacy-preserving synthetic data-generation models have been developed. These model are different categories that classify different types of models. 

First, there are models based on Generative Adversarial Networks (GANs), which use of a generator and a discriminator. The generator creates synthetic data that approximates the distribution of real data using Gaussian noise as input, while the discriminator identifies which data is synthetic and which is not. After a period of training, the model can generalize the structure of the synthetic data. Within this category, different approaches exist, including WGAN \cite{weng2019gan}, CTGAN \cite{xu2019modeling}, PATEGAN \cite{jordon2018pate}, and PATECTGAN \cite{rosenblatt2020differentially}. 
Another category includes models based on machine learning, but not GANs. Examples of this category include Variational Autoencoders (VAEs) \cite{xu2019modeling}, which learn the data distribution in latent space through an encoder-decoder structure, the Classification and Regression Trees (CART) model \cite{breiman2017classification}, and Long Short Term Memory Networks (LSTM) \cite{santhanam2020context}. 

Finally, we have statistical models-based synthetic data generation techniques. Although the term statistical can be very general, here we can categorize models whose internal basis is rooted in Bayesian or frequentist statistical theorems. This category includes models based on Markov chains, such as the Variable Markov Model (VMM) \cite{ferrando2018generating,wang2021synthetic}, models based on Bayesian networks \cite{bao2021synthetic}, models that estimate densities using kernels \cite{tang2015kerneladasyn,harder2021dp}, and models based on the study of copulas. Copula-based models analyse the distribution structure of the data by estimating the correlations between variables. Some notable studies in this field include  \cite{restrepo2023nonparametric} and \cite{SDV}. 

However, the study of privacy in the generation of synthetic data from copula-based models still requires further research. Within this area, models such as those used in \cite{gambs2021growing,li2014differentially}, employ differential privacy to develop models that generate data with privacy guarantees. At the same time, the study of nonparametric models for synthetic data generation has been relatively understudied. One of the few studies \cite{restrepo2023nonparametric} develops an algorithm based on nonparametric copulas for data generation, creating a model that depends only on the data, and a hyperparameter.

In this work, we propose the study of the non-parametric copula model, extending its scope to support both categorical and numeric data simultaneously. Furthermore, we have enhanced the model to include robust privacy guarantees. The extended model has been rigorously evaluated and compared with existing models across various dimensions, including the critical dimension of privacy.

The outline of the work is as follows: In Section II we provide the Background of terms to be studied. Section III documents the methods of the article, describing the contributions, evaluation framework and implementation details. Section IV presents the results and discussion, considering to different questions to be answered. Finally, Section V offers the conclusions of the work.

\section{Background}
In this section, we review the fundamental concepts necessary for understanding this work, particularly the definition of differential privacy and copula-based models. We formalize the foundational model for synthetic data generation using non-parametric copulas, which serves as the basis for the development of our proposed methodology.

\subsection{Differential privacy}
Differential Privacy (DP) has become a standard mechanism for privacy protection, being adopted in commercial and governmental enterprises, as well as in the academic field, mainly because of its mathematical properties. Data generated from DP algorithms can latter be shared with untrusted parties or released to the public while ensuring strict privacy guarantees. 
    \begin{definition} [$(\epsilon, \delta)$ -Differential Privacy \cite{dwork2008differential}] A randomized mechanism $\mathcal{M}$ with range $\mathcal{R}$ is $(\epsilon, \delta)$-DP if
        $$P[\mathcal{M(D) \in \mathcal{O}}] \leq e^{\epsilon} \cdot P[\mathcal{M(D') \in \mathcal{O}}] + \delta$$
        holds for any subset of outputs $\mathcal{O} \subseteq \mathcal{R}$ and for any adjacent datasets $\mathcal{D}$ and $\mathcal{D'}$, where $\mathcal{D}$ and $\mathcal{D}'$ differ from each other with only one training example, $\epsilon$ is the upper bound of privacy loss, and $\delta$ is the probability of breaching DP constraints.
    \end{definition}

    Typically, $\mathcal{M}$ denotes the training (generative) algorithm of a generative model, where DP ensures that the presence of an individual in the dataset remains difficult to detect. DP exhibits several key properties, including the post-processing property and the composition property.
    \begin{definition} [Post-processing \cite{dwork2014algorithmic}] If $\mathcal{M}$ satisfies $(\epsilon, \delta)$-DP, $F \circ \mathcal{M}$ will satisfy $(\epsilon, \delta)$-DP for any data-independent function $F$ with $\circ$ denoting the composition operator.
    \end{definition}

    \begin{definition}[Composition \cite{dwork2014algorithmic}] For every $\epsilon \geq 0$, $\delta \in [0,1]$, if $\mathcal(M)_{0}, \cdots, \mathcal(M)_{k-1}$ are each $(\epsilon, \delta)$-DP, then their composition $\mathcal(M)_{0} \circ \cdots  \circ \mathcal(M)_{k-1}$ is $(k \epsilon, k \delta)$-differentially private. 
    \end{definition}  

\subsection{Gaussian Copula}
        \label{sec: Gaussian Copula}
         Understanding the mathematical intricacies of this model is crucial, as it not only forms the foundation for many of the developments proposed in this paper, but also serves as a classical and transparent framework for comprehending the statistical structure of the data while preserving its distributional properties.

        Let $(X_{1},...X_{m})$ be a dataset and a Cumulative distribution Function (CDF) such as
        $$F_{i}=P(X_{i}\leq x)$$
        Consider the vector
        \begin{equation}
            (U_{i},\cdots, U_{m}) = (F_{1}(X_{1}),\cdots,F_{m}(X_{m}))
            \label{eq: Uniform_Dist}
        \end{equation}

        which means that the vector of cumulative distribution functions (CDFs) can be represented with uniform margins due to the application of the probability integral transform to each component.

        \begin{definition}[Copula and Sklar's Theorem \cite{nelsen2006introduction}]
            A m-dimensional copula $C:[0,1]^{m} \rightarrow [0,1]$ of a random vector $(X_{1},\cdots , X_{m})$ is defined as the joint distribution function (CDF) of $(U_{1}, \cdots, U_{m})$ on the unit cube $[0,1]^{m}$ with uniform margins
            $$C(u_{1}, \cdots , u_{m})= P(U_{1} \leq u_{1}, \cdots , U_{m} \leq u_{m})$$
            where each $U_{i} = F_{i}$.

            by Sklar's theorem, we can state that there exists an m-dimensional copula $C$ on $[0,1]^{m}$ with $F(x_{1}, \cdots, x_{m}) = C(F_{1},\cdots,F_{m}) \forall x \in \mathbb{R}^{m}$.
            If $F_{1}, \cdots, F_{m}$ are all continuous, then $C$ is unique.
            Conversely, if $C$ is a m-dimensional copula and $F_{1}, \cdots, F_{m}$ are distribution functions, then $C(u_{1}, \cdots, u_{m})=F(F_{1}^{-1}(u_{1}), \cdots, F_{m}^{-1}(u_{m}))$ where $F_{i}^{-1}$ is the inverse marginal of CDF $F_{i}$.
        \end{definition}
The copula represents the dependence on the uniform distribution. Even if the data should be continuous to guarantee the continuity of margins, discrete data in a large domain can be considered continuous because the cumulative density functions do not have jumps, which ensures the continuity of margins \cite{li2014differentially}.

One of the most widely known and commonly used copulas in the context of synthetic data generation is the Gaussian copula. This is primarily due to its convergence properties in multivariate data, as well as the fact that many real-world high-dimensional datasets exhibit Gaussian dependence structures \cite{nelsen2006introduction}.

\begin{definition} [The Gaussian Copula \cite{bouye2000copulas}]
If $\rho$ is a symmetric and positive definite matrix with diag $\rho =1$ which represents the correlation. The joint cumulative multivariate normal distribution, with mean zero and covariance equal to $\rho$ is represented as $\Phi _{\rho}$. The Gaussian copula can be written as: 
$$ C_{\rho}^{Gauss} (\textbf{u})= \Phi_{\rho} (\Phi^{-1}(u_{1}), \cdots, \Phi ^{-1} (u_{m}))$$
where $\Phi ^{-1}$ is the inverse cumulative distribution of a standard normal.
If $F_{i}(x_{i})=u_{i}$ is a Gaussian CDF, is it possible to obtain the density of the Gaussian copula which is the Gaussian dependence part,

$$ c_{\rho} ^{Gauss}(\textbf{u}) = \dfrac{1}{\sqrt{|\rho|}} exp \left ( -\dfrac{1}{2} \zeta(\textbf{u}) ^{T} (\rho ^{-1} - \mathbb{I}) \zeta (\textbf{u})\right )$$
where $\zeta(\textbf{u})= \begin{pmatrix}
    \Phi ^{-1} (u_{1})  \\
     \vdots \\
    \Phi ^{-1} (u_{m}) 
    \end{pmatrix} $, $|\cdot |$ is the determinant, and $\mathbb{I} \in \mathbb{R}^{m\times m}$. Finally, a multivariate Gaussian density can be written as the Gaussian dependence and margins:
$$\Phi_{\rho}= c_{\rho} ^{Gauss}(\textbf{u}) \prod _{i=1}^{m} \dfrac {\phi (\Phi^{-1}(u_{i}))}{\sigma_{i}}$$
with $\phi$ as the marginal of the multivariate $\Phi_{p}$.
\end{definition}

The estimation of the copula in real-world applications is usually hard since the copula is unknown. So, considering observations
$(X_{1}^{i},X_{2}^{i}, \cdots, X_{m}^{i}), \quad i= 1,2,\cdots,n$
coming from a random vector $(X_{1},X_{2}, \cdots, X_{m})$ with continuous marginals, with true observations of the copula represented as
$$(U_{1}^{i},U_{2}^{i}, \cdots, U_{m}^{i})= (F_{1}(X_{1}^{i}),F_{2}(X_{2}^{i}),\cdots,F_{m}(X_{m}^{i})), \quad i=1, \cdots, n$$
one could calculate the marginal distributions of $F_{i}$ using the empirical distributions, to construct a pseudo-copula. The empirical distributions are defined as

\begin{equation}
F_{k}^{n}= \dfrac{1}{n} \sum_{i=1}^{n} \textbf{1}(X_{k}^{i}\leq x)
\label{eq: empirical_marginal_dist}
\end{equation}
Then, the pseudo-copulas observations are $(\widetilde{U_{1}} ^{i},\widetilde{U_{2}} ^{i}, \cdots, \widetilde{U_{m}} ^{i} )=(F_{1}^{n}(X_{1}^{i}),F_{2}^{n}(X_{2}^{i}), \cdots, F_{m}^{n}(X_{m}^{i})) \quad i=1, \cdots, n$ and the empirical copula is defined as:

\begin{equation}
C^{n}(u_{1},\cdots, u_{m}) = \dfrac{1}{n} \sum_{i=1}^{n} \textbf{1} (\widetilde{U_{1}} ^{i} \leq u_{1}, \cdots, \widetilde{U_{m}} ^{i} \leq u_{m})
\label{eq: empirical_copula}
\end{equation}

Finally, after estimating the pseudo-copula, the next step is to estimate the matrix $\rho$. Li et al. \cite{li2014differentially} propose two distinct methods for this estimation. The first method involves using Kendall's $\tau$ rank correlation, while the second utilizes maximum likelihood estimation, with the pseudo-copula data serving as input.

\begin{definition}[Kendall's $\tau$ rank correlation \cite{demarta2005t}]
Kendall's $\tau$ rank correlation is calculated as
$$\rho _{\tau} (X_{1},X_{2})= E \left [ sign(X_{1}-\widetilde{X}_{1})(X_{2}-\widetilde{X}_{2})\right ]$$
where $(\widetilde{X}_{1}, \widetilde{X}_{2})$ is the second independent pair with the same distribution as $(X_{1},X_{2})$. 
\end{definition}
For estimating the correlation matrix $\rho$ one can construct an empirical estimate of Kendall's $\tau$ for each bivariate margin of the copula. Considering that $\rho_{\tau}$ depends only on the copula C \cite{demarta2005t} given by:
$$\rho_{\tau} (X_{1},X_{2}) = 4 \int_{0}^{1} \int_{0}^{1} C(u_{1},u_{2}) dC(u_{1},u_{2})-1$$
so, using the form of the Gauss copula $C_{\rho}^{Gauss}$, it is possible to get
$$\rho _{\tau}(X_{1},X_{2})= \dfrac{2}{\pi} \arcsin{\rho}$$
Using this result it is possible to infer an estimated version of $\rho$, such as

\begin{equation}
    \hat{\rho_{\tau}}(X_{j},X_{k})=\binom{n}{2}^{-1} \sum_{1 \leq i_{1} < i_{2} \leq n} sign(X_{i_{1},j}-X_{i_{2},j})(X_{i_{1},k}-X_{i_{2},k})
    \label{eq: rho_est}
\end{equation}

getting an unbiased and consistent estimator, with $n$ as the number of samples in $X_{j}$. To obtain the estimator for the entire matrix $\rho$, it could be possible to define an empirical Kendall's $\tau$ matrix $R^{\tau}$, defined by $R_{jk}^{\tau} = \hat{\rho_{\tau}}(X_{j},X_{k})$ and build the estimator $\hat{\rho}=\sin{\left ( \dfrac{\pi}{2} R^{\tau}\right )}$. Since there is no guarantee the matrix is positive definite, it can be adjusted using any procedure \cite{demarta2005t}. The algorithm \ref{algorithm: GaussianSDG} shows how to sample synthetic data with Gaussian dependency. 
\begin{algorithm}[H]
        \scriptsize
         \caption{Sampling data from Gaussian Copula}
         \label{algorithm: GaussianSDG}
        \begin{algorithmic}[1]        
        \State \textbf{Input:} Marginal distributions, correlation matrix $\rho$
        \State \textbf{Output:} Synthetic data
        \State
        \State Generate pseudo-copula synthetic data $(\hat{T_1}, \ldots, \hat{T_m})$:
        \State \hspace{\algorithmicindent} \textbf{a.} Generate a multivariate random number vector $(\hat{X_1}, \ldots, \hat{X_n})$ \\
        \hspace{\algorithmicindent} following the Gaussian joint distribution $\Phi_{\rho}$.
        \State \hspace{\algorithmicindent} \textbf{b.} Transform \textbf{$(X_1, \ldots, X_m)$} to $(\hat{T_1}, \ldots, \hat{T_m})$, using $\hat{T_{j}}=\phi(\hat{X_{j}}), \quad j=1, \ldots,m$ with $\phi(\hat{X_{j}})$ is the standard Gaussian distribution.
        \State
        \State Compute synthetic data $D$ as follows:
                $$\hat{D}=(F_{1}^{-1}(\hat{T_{1}}), \ldots, F_{m}^{-1}(\hat{T_{m}}))$$
                with $F_{j}^{j}(\hat{T_{j}})$ as the inverse of the empirical marginal distribution function.
        \end{algorithmic}
        \end{algorithm}

\subsection{Differentially Private Copula (DPCopula)}
Starting from the general framework with a Gaussian Copula in Section \ref{sec: Gaussian Copula}, we can see that the data is only accessed in two different sections: When the marginals are generated, and when the correlation matrix $\rho$ is calculated. Li et al. \cite{li2014differentially} build a process to generate synthetic data with Differential Privacy using Gaussian copula. They use a DP histogram to obtain the marginal distributions and injected Laplacian noise in the two implemented methods for finding the $\rho$ matrix.

One could implement DP to a histogram with a naive solution. Given an attribute $X$ with the value set $\mathcal{V}$ in a database $\mathcal{D}$, build a frequency vector of size $|\mathcal{V}|$ with the $i^{th}$ as the number of tuples $t \in \mathcal{D}$, with $t \cdot X=v_{i} \in \mathcal{V}$. A histogram $H$ over the attribute $X$ is built when a frequency vector is partitioned into a set of bins $\{H_{1}, \cdots, H_{n}\}$, where each value $H_{j}$ specifies a range of values it covers, and assigns each value a representative count. 
The bins are non-overlaping intervals of the attribute and satisfy the condition $|\mathcal{D}|= \sum_{i=1}^{n}H_i$. 
 For a histogram $H$ with bins $\{H_{1}, \cdots, H_{n}\}$, the private version will be
$$\hat{H}=\left \{H_{1}+\mathcal{L}\left ( \frac{1}{\epsilon}  \right ),\cdots,H_{n}+\mathcal{L}\left ( \frac{1}{\epsilon}  \right )\right \} $$
More efficient methods for this calculation exist, as excessive noise may be introduced to the data, resulting in a loss of information and utility. Acs et al. \cite{acs2012differentially} introduced a Fourier Perturbation Algorithm, known as EFPA, which applies the Fourier transform to a histogram and compresses it by removing high-frequency components using the exponential mechanism. Following a similar approach to the Basic Fourier Perturbation Algorithm, EFPA is presented in Algorithm \ref{algorithm: EFPA}.
\begin{algorithm}[H]
        \scriptsize
         \caption{Enhanced Fourier Perturbation with DP \cite{acs2012differentially}}
          \label{algorithm: EFPA}
        \begin{algorithmic}[1]
        \State \textbf{Input:} Histogram $H$ with length n, where n is odd
        \State \textbf{Input:} Privacy budget $\epsilon$
        \State \textbf{Output:} Noisy histogram $\hat{H}$
        \State
        \State Compute the DFT coefficients $\textbf{F} := DFT^{real} (H)$
        \State Select the number of coefficients to operate $m:= \frac{(n+1)}{2}$
        \State Compute utility function $u(H,k)=\sqrt{\sum _{i=k+1}^{m} 2 |F_{i-1}|^{2}} + \frac{2z}{\epsilon}$ for all $1\leq k \leq m$, where $z=2k+1$
        \State Select $k$ with exponential mechanism $\propto exp \left (  - \frac{\epsilon \cdot u(H,k)}{4}\right )$
        \State Recalculate $z:= 2k+1$
        \State $\hat{\textbf{F}}^{k}:= \textbf{F}^{k} + \left\langle  \mathcal{L}(2\sqrt{z}/\epsilon)\right\rangle ^{k}$ where $\textbf{F}^{k}$ denotes the first $k$ elements of $\textbf{F}$
        \State Pad $\hat{\textbf{F}}^{k}$ to be $n$-dimensional, appendind $n-k$ zeros, denoted as $PAD^{n}(\hat{\textbf{F}}^{k}$
        \State $\hat{H}=IDTF(PAD^{n}(\hat{\textbf{F}}^{k}))$
        
        \end{algorithmic}
        \end{algorithm}

        Finally for computing DP correlation matrix estimator, is it possible to use equation \ref{eq: rho_est} for which the transformation will result as
        
        $$
                    \hat{\rho_{\tau}}(X_{j},X_{k})=\binom{n}{2}^{-1} \sum_{1 \leq i_{1} < i_{2} \leq n} sign(X_{i_{1},j}-X_{i_{2},j})(X_{i_{1},k}-X_{i_{2},k}) + \mathcal{L}\left (\frac{ \binom{m}{1} \Delta}{\epsilon} \right )
        $$

        where $\Delta$ is the sensitivity of each pairwise Kendall's $\tau$ coefficient with a value of $\frac{4}{n+1}$. The proof can be found at \cite{li2014differentially}.

\subsection{Non Parametric Copula (NPC)}
\label{sec: NPC}
This method is formulated by Restrepo et al. \cite{restrepo2023nonparametric}. Considering equation \ref{eq: Uniform_Dist}, one can say that since $F_{j}$ is a non-decreasing function, with the random vectors $[U_{1}, \cdots, U_{m} ]$ and $[X_{1}, \cdots, X_{m}] =[F_{1}(X_{1}), \cdots, F_{p}(X_{p})]$ there is a procedure to generate, from a known copula $C$, observations of the random vector $[U_{1}, \cdots , U_{m}]$ to obtain a sample
$[X_{1}, \cdots, X_{m}]$ with $[F_{1}^{-1}(U_{1}), \cdots , F_{m}^{-1}(U_{m})]$. However, this is only possible if both, the Copula and the Empirical Distributions are known. In equation \ref{eq: empirical_marginal_dist} we already introduced a way to generate empirical marginal distributions. Let us consider a dataset $X \in \mathbb{R}^{n \times m}$. It is possible to define a empirical copula as the empirical distribution of the rank transformed data, rewritting equation \ref{eq: empirical_copula} as
$$ \hat{U}_{j,i}= \frac{1}{n} \sum _{k=1}^{n} \textbf{1} (X_{ki} \leq X_{ji}), \quad \forall i \in [1, \cdots, m]$$
Those components can also be written as $\hat{U}_{j,i}= R_{j,i} / n$, which represents the rank of the observation $X_{ji}$. This procedure only consider the m-dimensional support of the empirical copula estimated, so Restrepo et al. \cite{restrepo2023nonparametric} introduced a natural estimator of $F_{i}$. Consider a partition of the $i^{th}$ column in $X$ of the interval $[X_{[1]i},X_{[n],i}]$ as $X_{[1]i} = a_{0i}<a_{1i}< \cdots < a_{t_{i}i}=X_{[n]}$. Here, $X_{[r]i}$ represents the $r^{th}$ order statistic of a random sample $X_{1i},\cdots, X_{ni}$, that is $$X_{[1]i}\leq X_{[2]i} \cdots\leq X_{[n]i}$$ $B_{si}$ is defined as
$$B_{si}= \left\{\begin{matrix}
 [a_{s-1},a_{s}] & if s=1\\
 (a_{s_1},a_{s}] & otherwise

\end{matrix}\right.$$

$$R(B_{s})= \frac{1}{n} \sum_{j=1}^{s} \sum_{k=1}^{n} \textbf{1}(x_{ki}\in B_{ji}) \quad \forall s \in \{ 1, \cdots, t_{i}\}$$
They showed that $R(B_{si})$ is a natural unbiased estimator of $F_{i}(a_{si})$. With $d$ as a value generated from a discrete uniform distribution in $\{1,\cdots, n\}$, they selected the $d^{th}$ row of the m-dimensional support of the empirical copula previously calculated. Then, by generating a random variable $U  \sim \text{Uniform}[0,1]$, and for each $i \in \{1, \cdots, m\}$ get $min\{s | R(B_{si}) \geq \hat{U}_{d,i}\}$, it is possible to generate synthetic data $[\hat{X}_{1}, \cdots, \hat{X}_{m}]$ considering 
$$\hat{X}_{i} = a_{(s-1)i}+(a_{si}-a_{(s-1)i})U \quad \forall i \in \{1, \cdots, m\}$$
The complete step-by-step algorithm can be found in \cite{restrepo2023nonparametric}.

\section{Methods}
In this section, we describe implementation of the Differentially Private Non-Parametric Copula method, an extension of NPC method with privacy guaranties. Also, we outline the evaluation framework, and then provide details regarding the implementation. 
\subsection{Differentially Private Non-Parametric Copula (DPNPC)}
 \label{sec: DPNPC}
 The method originally formulated by Restrepo et al. \cite{restrepo2023nonparametric} lacks inherent privacy-preserving mechanisms. Leveraging the structure proposed by \cite{li2014differentially}, we construct an approximation of the Nonparametric Copula (NPC) method using Differential Privacy (DP) as the privacy-preserving mechanism. This modified approach is now referred to as DPNPC. An important observation regarding the original NPC method is that the data serves two primary purposes: (I) to generate the empirical distribution function for the $i^{th}$ variable at $X$, and (II) to generate the frequency tables with $T[i]$ bins for the $i^{th}$ variable in $X$. 
The original NPC synthetic data generation algorithm is included as algorithm \ref{algorithm: NPC}.

\begin{algorithm}[H]
        \scriptsize
        \caption{NPC (\cite{restrepo2023nonparametric})}
        \label{algorithm: NPC}
        \begin{algorithmic}[1]
        
        \State \textbf{Input:} \\
        $X \leftarrow \mathbb{R}^{n \times p}$ matrix of the real data
        \\
        $N \leftarrow$ number of synthetic observations to generate
        \\
        $T \leftarrow \mathbb{R}^{p}$ selected number of bins
        \State Initialize $U$ as an array of zeros size $n \times p$
        \State Initialize $Y$ as an array of zeros of size $N \times p$
        \State Initialize $D$ as a list of size $N$ filled with randomly selected integers between $1$ and $n$
        \For{$i \leftarrow 1 \text{ to } p$}
        \State \textcolor{red}{Generate the empirical distribution for $i^{th}$ variable}
        \State \textcolor{red}{Generate the frequency tables with $T[i]$ bins for the $i^{th}$ variable}
        \EndFor
        \State Initialize a counter variable \textit{count} as zero
        \For{$i \leftarrow 1 \text{ to } p$}
            \For{$j \leftarrow 1 \text{ to } n$}
                \State $U[i,j] \leftarrow$ the empirical distribution function value for $X[j,i]$
            \EndFor
        \EndFor
        \For{$d \in D$}
        \State Initialize $K$ as an array of zeros of size $1 \times p$
        \For{$i \leftarrow 1 \text{ to } p$}
        \State Find the corresponding class interval for $U[d,i]$ in the respective frequency table for $i^{th}$ column
        \State Generate a uniformly distributed number in the corresponding class interval of $U[d,i]$
        \State Store the generated number in $K[1,i]$
        \EndFor
        \State Replace row number \textit{count} in $Y$ for $K$
        \State \textit{count} $\leftarrow$ \textit{count} +1
        \EndFor
        
        \end{algorithmic}
        \end{algorithm}

Within the NPC algorithm \ref{algorithm: NPC} steps highlighted in red, it is possible to identify a code fragment that directly accesses the data, which is where a privacy break might occur when accessing to the original data. Here, we employ the EFPA algorithm to generate differentially private histograms to ensure privacy. The privacy budget is evenly divided, allowing us to generate empirical marginals through differentially private observations, and subsequently, to generate the $U$ matrix of frequencies through another histogram made with the number of bits that acts as a parameter in the method. Following this approach, the NPC algorithm is updated to become the DPNPC method as shown in algorithm \ref{algorithm: DPNPC}.

\begin{algorithm}[h]
    \scriptsize
         \caption{DPNPC}
         \label{algorithm: DPNPC}
        \begin{algorithmic}[1]
        
        \State \textbf{Input:} \\
        $X \leftarrow \mathbb{R}^{n \times p}$ matrix of the real data
        \\
        $N \leftarrow$ number of synthetic observations to generate
        \\
        $T \leftarrow \mathbb{R}^{p}$ selected number of bins
        \\
         \textcolor{green}{$\epsilon \leftarrow$ privacy budget to spent}
        
        \State Initialize $U$ as an array of zeros size $n \times p$
        \State Initialize $Y$ as an array of zeros of size $N \times p$
        \State Initialize $D$ as a list of size $N$ filled with randomly selected integers between $1$ and $n$
        
        \For{ $i \leftarrow 1 \text{ to } p$}
         \State \textcolor{green}{Get the unique values of attribute $i^{th}$}
         \State \textcolor{green}{Get the marginal histogram for $i^{th}$ for every unique value}
         \State \textcolor{green}{Inject noise using EFPA algorithm, using a privacy budget of $\epsilon/(2p)$ to get the $i^{th}$ DP marginal}
         \State \textcolor{green}{Get the empirical cumulative distribution function given the $i^{th}$ DP marginal distribution }
        \EndFor

        \For{$i \leftarrow 1 \text{ to } p$}
        \State \textcolor{green}{Build a DP frequency table by building a histogram with a selected number of pins $T[i]$}
        \State \textcolor{green}{Inject noise using EFPA algorithm, spending a privacy budget of $\epsilon/(2p)$}
        \EndFor
             
        \State Initialize a counter variable \textit{count} as zero
        \For{$i \leftarrow 1 \text{ to } p$}
            \For{$j \leftarrow 1 \text{ to } n$}
                \State \textcolor{green}{$U[i,j] \leftarrow$ the DP empirical distribution function value for $X[j,i]$}
            \EndFor
        \EndFor
        
        \For{$d \in D$}
        \State Initialize $K$ as an array of zeros of size $1 \times p$
        \For{$i \leftarrow 1 \text{ to } p$}
        \State Find the corresponding class interval for $U[d,i]$ in the respective frequency table for $i^{th}$ column
        \State Generate a uniformly distributed number in the corresponding class interval of $U[d,i]$
        \State Store the generated number in $K[1,i]$
        \EndFor
        \State Replace row number \textit{count} in $Y$ for $K$
        \State \textit{count} $\leftarrow$ \textit{count} +1
        \EndFor
        
        \end{algorithmic}
        \end{algorithm}

The proposed DPNPC model \ref{algorithm: DPNPC} takes advantage of the properties of DP, using Sequential Composition to partition the privacy budget, and the post-processing property to elaborate the post-processing clusters after the construction of the DP marginal distributions and the frequency table. 

\subsection{Evaluation framework}
In this section, we are going to describe the process used at the evaluation phase of the pipeline, comparing different synthetic generation methods, using a set of metrics. 
\subsubsection{Preprocessing}
It is well-known that data needs to be transformed in various ways depending on the nature of different models. In this paper, we implement a treatment according to needs of model. In particular, we extended the NPC method to support categorical and numeric data using a encoding method. 
\begin{itemize}
        \item \textbf{DPNPC encoding}: 
    It was necessary to convert all categorical data into continuous values for this category. To achieve this, a Uniform-Encoder was implemented, based on the formulation from \cite{SDV}. The encoder replaces categorical values in the column with values in the range $[0,1]$. The Uniform-Encoder method has been included as algorithm \ref{algorithm: UniformEncoder}.

\begin{algorithm}[H]
         \scriptsize

         \caption{Uniform-Encoder \cite{SDV}}
         \label{algorithm: UniformEncoder}
        \begin{algorithmic}[1]
        
        \State \textbf{Input:} $X \leftarrow $ Categorical vector to be transformed
        \State Sort the categories from most frequent occurring to least
        \State Split the interval $[0,1]$ into sections based on the cumulative probability of each category
    
        \State Find the interval $[a,b] \in [0,1]$ that corresponds to the category according to the proportion of each of the categories. 
        \State Chose value between $a$ and $b$ by sampling from a truncated Gaussian distribution with $\mu$ as $(b-a)/2$ and $\sigma= (b-a)/6$   
        \State Generate a random number coming from the corresponding truncated distribution of the category in each value in $X$.
        \State Return the encoded variable $\hat{X}$
        \end{algorithmic}
        \end{algorithm}

     After the data sets are generated, the inverse transform is calculated by finding the interval that correspond to the category. 
\end{itemize}
Nan values from databases are eliminated in this step.

\subsubsection{Privacy Evaluation}

Attack-based privacy metrics focus on calculating the performance of an adversary, who aims to extract sensitive information from a dataset without authorization and measure the algorithm's efficiency according to its capacity to keep the data private. Inspired in the pipeline formulated by Giomi et al. \cite{giomi2022unified}, considering a framework for the attack, evaluating and estimating the risk of different datasets, we implemented a version of the Membership Inference Attack, and measure the performance using their risk calculation method. 
This attack happens whenever it is possible to link one original record to a set of records synthetically generated. For a collection of $N_{A}$ original records, the algorithm finds the k-closest synthetic records. Once this is calculated, the Gower distance  \cite{gower1971general} between the attacked record, and the closest neighbor is calculated, and the attack is considered successful if the distance is less than a tolerance.
The risk calculation consider three different attack phases:
 \begin{itemize}
     \item \textbf{Main:} In this phase, the synthetically generated dataset ( $X_{syn}$) is used to deduce private information of records in the training sample ($X_{train}$) i.e. the original dataset. 
     \item \textbf{Naive:} In this phase, a random guessing mechanism is used, to provide a baseline against which the strength of the main attack can be compared.
     \item \textbf{Control:} In this phase, a separate data set coming from the original, but that was not used to generate synthetic data is used to calculate a privacy risk. This measure helps us to distinguish the concrete privacy risk of the original data from the general risk intrinsic to the whole population. 
 \end{itemize}

 The three phases generate a set of guesses $g=\{g_{1}, \cdots, g_{N_{A}}\}$ on $N_{A}$ target records. Then, an evaluation phase starts, comparing the guesses versus the truth of the data, generating a vector $o=\{o_{1}, \cdots, o_{N_{A}}\}$, where $o_{i}=1$ if the $i^{th}$ guess $g_{i}$ is correct. To measure the Risk, a quantification phase rates the success of the privacy attack from the evaluation with a measure of statistical uncertainties. Assuming the outcome $o_{i}$ of each attack follows a Bernoulli trial distribution, the true privacy risk $\hat{r}$ can be calculated with an estimation considering a confidence interval $\hat{r} \in r \pm \delta_{risk}$. \cite{giomi2022unified} calculated the risk factor using a confidence level $\alpha$ via the Wilson Score Interval
 $$
 r= \dfrac{N_{S}+z_{\alpha}^{2}/2}{N_{A}+z_{\alpha}^{2}}
 $$
 $$ \delta_{risk}= \dfrac{z_{\alpha}}{N_{A}+z_{\alpha}^{2}} \sqrt{\dfrac{N_{S}(N_{A}-N_{S})}{N_{A}}+\dfrac{z_{\alpha}^{2}}{4}}$$

 with $N_{S} = \sum _{i=1}^{N_{A}} o_{i}$, and $z_{\alpha}$ the inverse of the cumulative distribution function of the normal distribution. The risk rates are calculated for the \textit{main}, \textit{naive}, and \textit{control} attacks as $(r_{train} \pm \delta_{train})$, $(r_{naive} \pm \delta_{naive})$, and $(r_{control} \pm \delta_{control})$. An attack is considered as successful if $r_{naive}<r$, which means that the attack was stronger than the naive baseline. Finally, a risk $R$ is calculated considering the \textit{control} attack, derived as:
 $$R = \dfrac{r_{train}-r_{control}}{1- r_{control}}$$
 $R$ measures, on the numerator the excess of attacker success, and the denominator the maximum improvement over the control attack. 

\subsubsection{Utility Evaluation}
The utility method used to evaluate the performance of the models involves implementing a binary classifier, specifically XGBoost. This approach compares the results of a classifier trained on synthetic data with those trained on real data for a particular attribute to be predicted. Ultimately, the models are tested on a separate test dataset that was not used for training either the classifier models or the data-generating model.

Ideally, a classifier trained on synthetic data should exhibit classification performance comparable to that of one trained on real data. This comparison is conducted using the Matthews Correlation Coefficient (MCC)\cite{matthews1975comparison}, formally defined as: 

\[
MCC= \frac{\frac{TP}{N}- (S \times P)}{\sqrt{ (S \times P) (1-S) (1-P)}}
\]
where $N =$ Number of records, $TP =$ True positive rate, $FN =$ False negative rate, $FP =$ False positive rate, $S = \frac{TP + FN}{N}$ and $P= \frac{TP+FP}{N}$. The measure is between -1 and 1, such that 1 would imply a perfect classifier.

\subsubsection{Fidelity Evaluation}
As a fidelity metric, we use the Kolmogorov-Smirnov distance to assess how closely the distributions of the synthetically generated data approximate those of the original data.The KS distance, which ranges from 0 to 1, is calculated for each attribute, and the average distance is reported for each of the generated datasets. This test evaluates the hypothesis that the reference and experimental distributions follow the same distributional law, being considered valid only if the test statistic $D_{KS}$ is close to a threshold $\delta(\alpha)$. 

We consider a reference distribution $f_{t}$ and an experimental distribution $f_{e}$, along with their cumulative distribution functions $F_{t}$ and $F_{e}$. The statistical test is formally expressed as follows:
\[
D_{KS}(F_{t},F_{e})= \underset{x}{sup} \left | F_{t}(x) - F_{e}(x) \right |
\]

\subsection{Implementation details}

We compare our method using the pipeline developed by Gambs et al. \cite{gambs2021growing}, evaluating it against three additional models (PrivBayes, DP-Copula, DP-Histogram) to verify the previously mentioned metrics. We compare the implementation of a naively differentially private histogram (DP-Histogram), which adds Laplacian noise with a mean of 0 and a scale of $\frac{\Delta}{\epsilon}$, where $\Delta = 2$, to each bin count in the histogram. 

Additionally, we use the PrivBayes implementation provided by \cite{gambs2021growing} and referenced by \cite{bowen2019comparative}, running experiments with a chosen $\epsilon$ parameter and a maximal number of parent nodes in the Bayesian Network set to 3. Finally, we compare our implementation with the DP-Copula model \cite{li2014differentially}, which, as implemented by \cite{gambs2021growing}, uses a parameter to allocate the privacy budget between the computation of marginal densities and the generation of the correlation matrix. In this case, the parameter was set by default, with half of the privacy budget dedicated to each process. Similarly, the \textit{bins} parameter associated with DPNPC was fixed at 40.
\subsubsection{Datasets}
In order to compare our results with the reference paper by Gambs et al. \cite{gambs2021growing}, we used three different public dataset, which contains various dimensions and attribute types. The first one is Adult Dataset from UCI \cite{dua2017uci}, with 32 561 profiles, 8 categorical values, and 6 discrete values. The second is the COMPAS dataset  \cite{angwin2019machine}, with 10 568 registers, with 13 attributes, and finally the Texas Hospital dataset \cite{Texas2013}, a sample of 150 000 from a original dataset with 636 140 records, and 17 attributes, from which 11 are categorical. 
\subsubsection{Parameters for data metrics}

Regarding the evaluation of the privacy metric, the $\delta$ associated with the tolerance of the metric was set to $0.10$. Additionally, for the \textit{Adult} and \textit{Compas} datasets, a total of 250 attacks were executed, while for the \textit{Texas Hospital} dataset, 1000 attacks were conducted.

For generating the utility metric, binary classification was performed on the following attributes: \textbf{salary} for \textit{Adult}, \textbf{is violent recid} for \textit{Compas}, and \textbf{ethnicity} for \textit{Texas Hospital}. Each experiment involved generating datasets with varying $\epsilon$ parameters within the range $\epsilon \in \{0.1, 0.2, 0.3, 0.4, 0.5, 0.8,1.0\}$ $\cup$
$\{ 2.0, 5.0, 10.0, 15.0\}$ to understand the behavior of the models under different levels of privacy protection, and for privacy, generated data with $\epsilon = 0$

Furthermore, to solve question Q1 \ref{sec: Q1}, we iterated over the $\epsilon$ parameter and the hyperparameter \textit{bins} for each dataset, with \textit{bins} varying in the range $[10, 100]$.

\section{Results and discussion}
Given the nature of the method, which generates synthetic data using a uniform kernel that follows the correlations of the pseudo-copula, calculating the probability that a generated data point exactly matches one of the training data points could provide insights into the privacy of the method.

Therefore, it is essential to determine how the resolution of the generated grid, based on the number of bins, affects the privacy, utility, and similarity of the synthetic data. With this objective, the following questions are proposed for testing:

\subsection{Q1: Is it better to add noise via DP, or make resolution lower for privacy porpoises?}
\label{sec: Q1}
It is evident that while the non-private NPC version does not offer the indistinguishability benefits provided by Differential Privacy (DP), the resolution of the method influences how close the data are generated respect to the original distribution. This ensures that, in some manner, data are generated according to the multivariate distribution defined by the pseudo-copula, without necessarily adhering to a distance metric that would require the generated data to be exactly identical to the training data.

\begin{figure}[htpb]
    \centering
    \begin{subfigure}[b]{\linewidth}
        \centering
        \includegraphics[width=\linewidth]{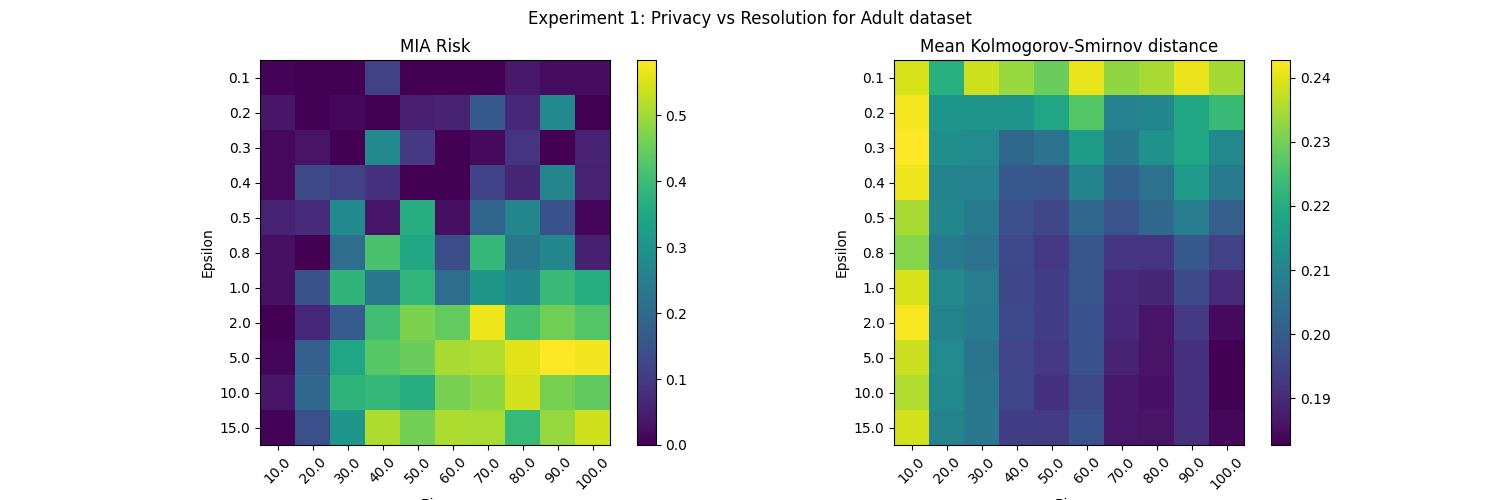}
      
        \label{fig:PvsR_adult}
    \end{subfigure}
    
    \begin{subfigure}[b]{\linewidth}
        \centering
        \includegraphics[width=\linewidth]{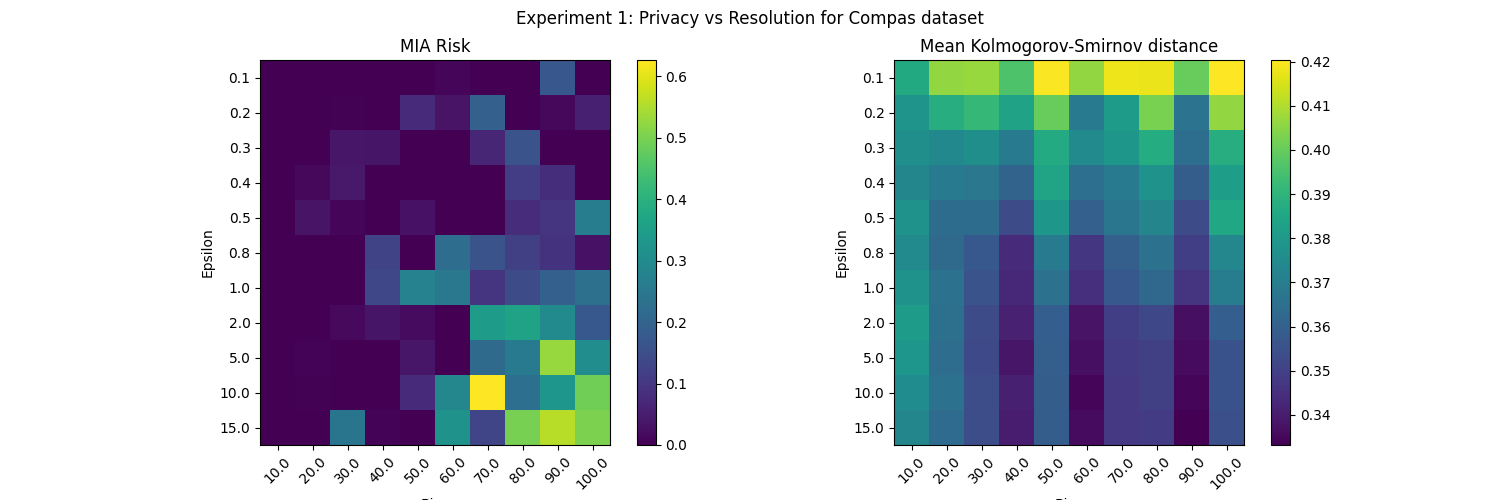}
       
        \label{fig:PvsR_compas}
    \end{subfigure}
    
    \begin{subfigure}[b]{\linewidth}
        \centering
        \includegraphics[width=\linewidth]{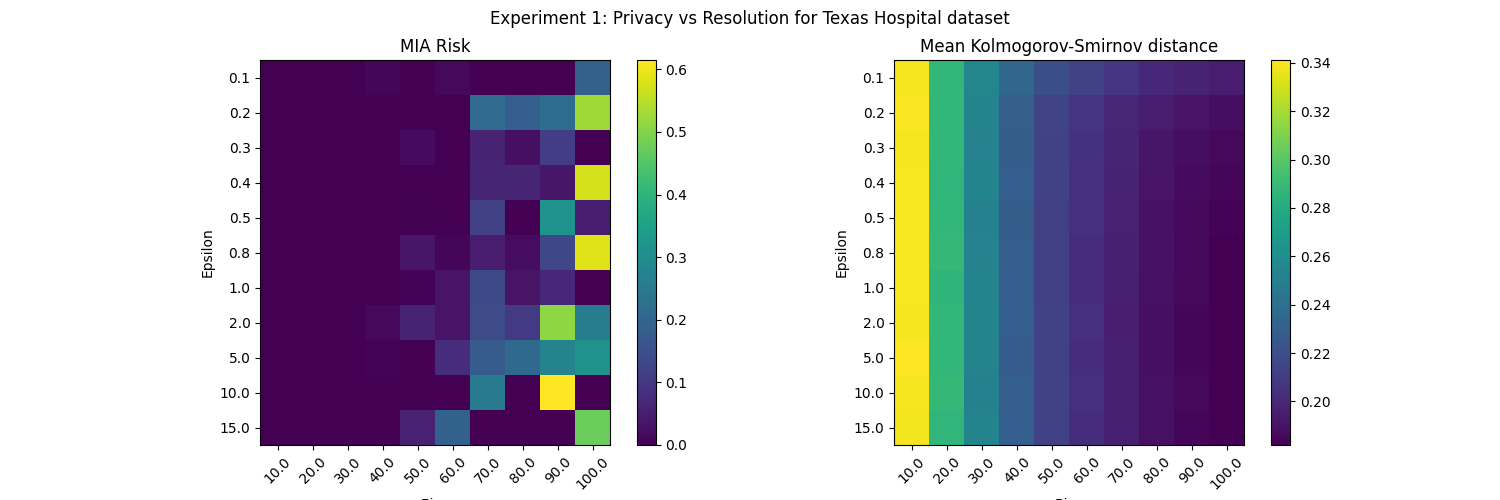}
        
        \label{fig:PvsR_texas}
    \end{subfigure}
    
    \caption{Privacy and distance measures versus bins for different datasets using the DPNPC model. The first plot represents the behavior for the \textit{Adult} dataset, the second for the \textit{Compas} dataset, and the third for the \textit{Texas Hospital} dataset. The y-axis corresponds to the $\epsilon$ values, while the x-axis indicates the bins values.}
    \label{fig:PvsR_combined}
\end{figure}

The results presented in Figure \ref{fig:PvsR_combined} suggest a stable behavior of the algorithm. It is observed that as the privacy parameter $\epsilon$ and the resolution parameter \textit{bins} increase, the success rate of attacks using MIA rises. Conversely, when these parameters are smaller, the distance tends to be greater. These results are consistent across the three datasets, with specific combinations of parameters, particularly for \textit{bins}, either benefiting or impairing the metric outcome due to the nature of the data. This variability is likely due to the sensitivity of certain attributes' distributions to the number of \textit{bins} used in training the model.

\subsection{Q2: Is data best modelled with DPNPC method instead of other statistical methods?}
To validate the model and compare it with similar methods, we evaluated the privacy, utility, and fidelity metrics of DPNPC against other methods described in the literature, measuring its performance across different values of $\epsilon$.

\subsubsection{Utility}
\begin{figure*}[h]
    \centering
    \begin{subfigure}[b]{0.45\linewidth}
        \centering
        \includegraphics[width=\linewidth]{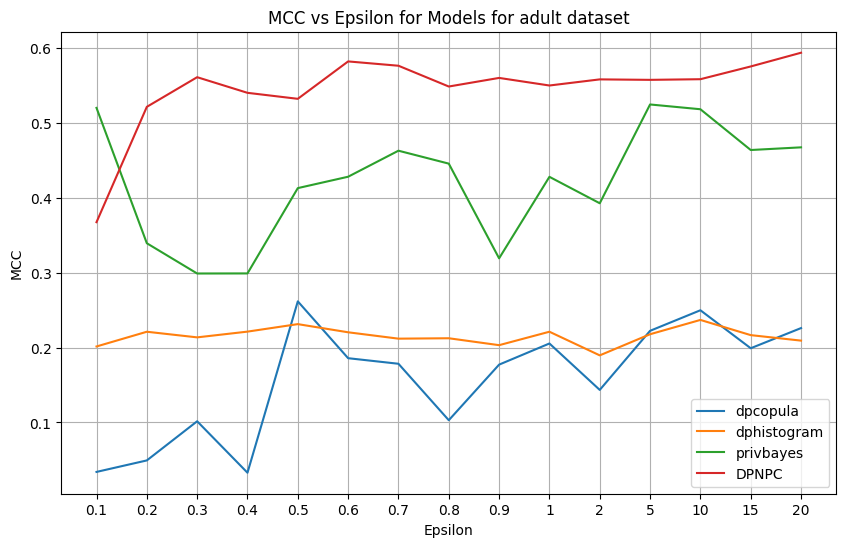}
        \caption{}
        \label{fig:utility_adult}
    \end{subfigure}
    \hfill
    \begin{subfigure}[b]{0.45\linewidth}
        \centering
        \includegraphics[width=\linewidth]{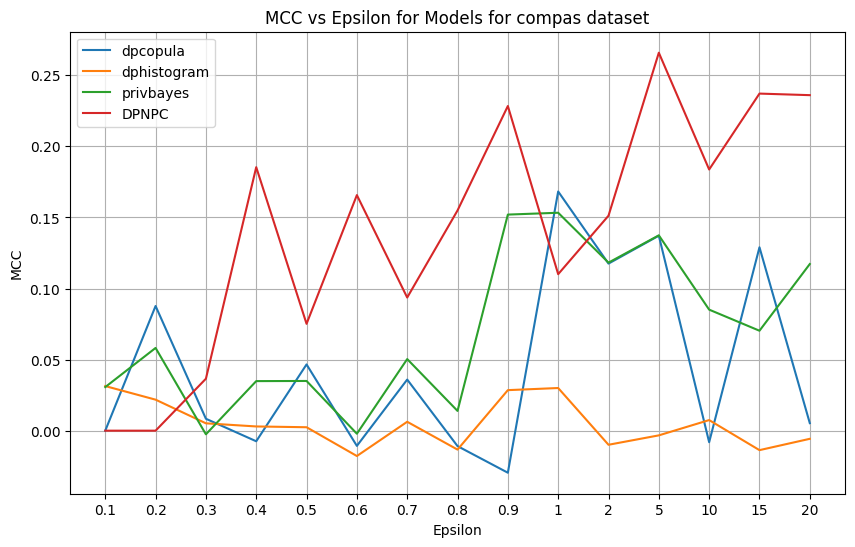}
        \caption{}
        \label{fig:utility_compas}
    \end{subfigure}
    
    \vspace{1cm}
    \begin{subfigure}[b]{0.45\linewidth}
        \centering
        \includegraphics[width=\linewidth]{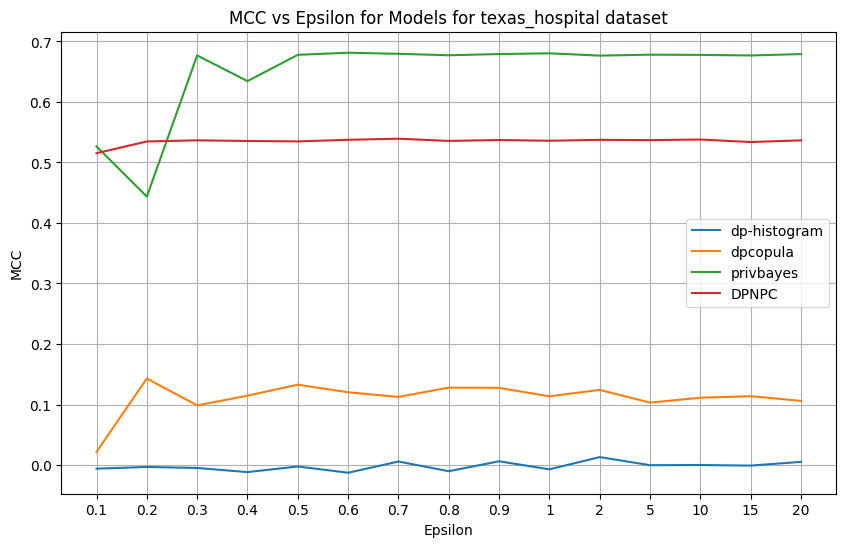}
        \caption{}
        \label{fig:utility_texas}
    \end{subfigure}

    \caption{Utility measures (a)\textit{Adult}, (b)\textit{Compas} and (c)\textit{Texas Hospital} dataset, with the MCC metric on the y-axis and different $\epsilon$ values along the x-axis.The closer the metric is to 1, the better the utility}
    \label{fig:utility_combined}
\end{figure*}

The analysis using the Matthews Correlation Coefficient (MCC) as the utility metric provides the results for the three databases, as depicted in Figure \ref{fig:utility_combined}. Notably, the DPNPC and PrivBayes methods demonstrate the best performance in maintaining utility properties across different $\epsilon$ values for all three databases. In large databases, these models maintain their properties with relatively high utility values, while the smallest database shows significant variability in classification model performance. For databases with a high number of categorical variables, such as the Texas Hospital dataset, PrivBayes performs notably well for $\epsilon$ values greater than $0.3$. 

In contrast, for databases with a more balanced distribution of categorical and continuous attributes, DPNPC shows superior performance. This may be attributed to the effects of encoding on the sample structure when implementing DPNPC. Models such as DP-Histogram and DP-Copula, according to this metric, exhibit the poorest performance in preserving the multivariate dependence structure, highlighting their inferior performance.

\subsubsection{Fidelity}
\begin{figure*}[htb]
    \centering
    \begin{subfigure}[b]{0.45\linewidth}
        \centering
        \includegraphics[width=\linewidth]{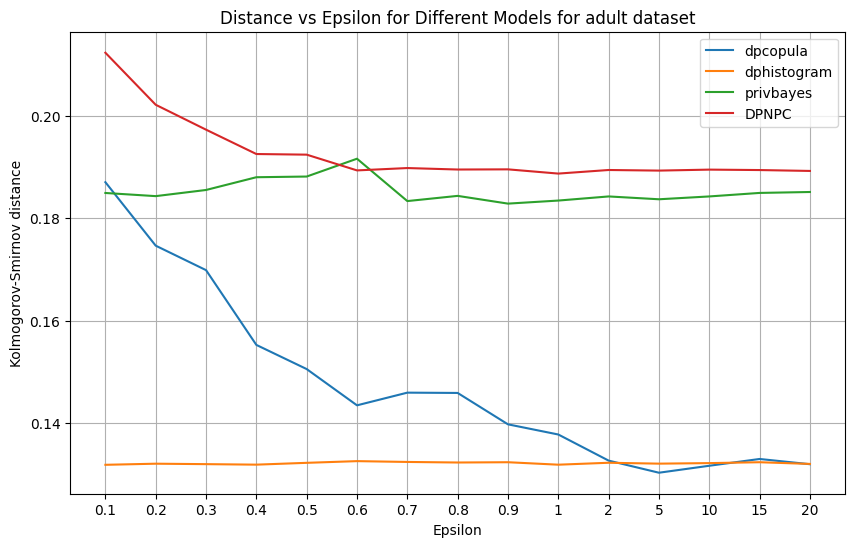}
        \caption{}
        \label{fig:fidelity_adult}
    \end{subfigure}
    \hfill
    \begin{subfigure}[b]{0.45\linewidth}
        \centering
        \includegraphics[width=\linewidth]{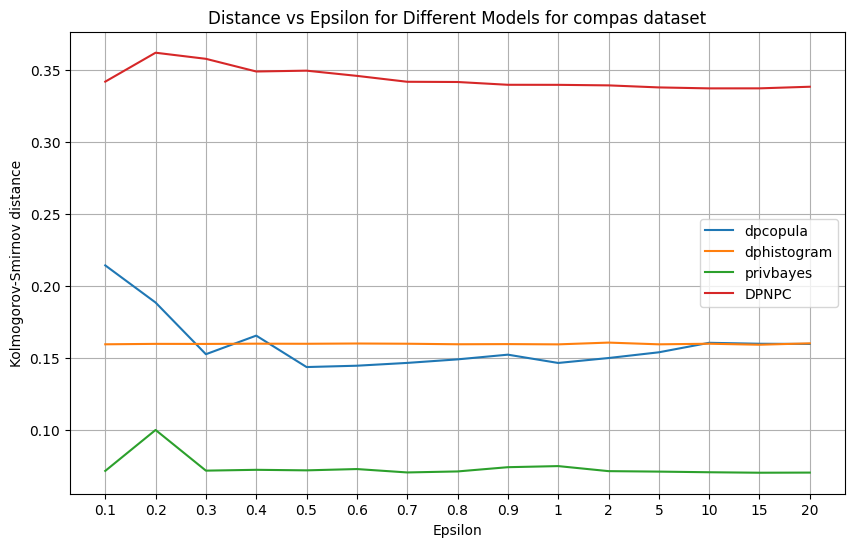}
        \caption{}
        \label{fig:fidelity_compas}
    \end{subfigure}
    
    \vspace{1cm}
    \begin{subfigure}[b]{0.45\linewidth}
        \centering
        \includegraphics[width=\linewidth]{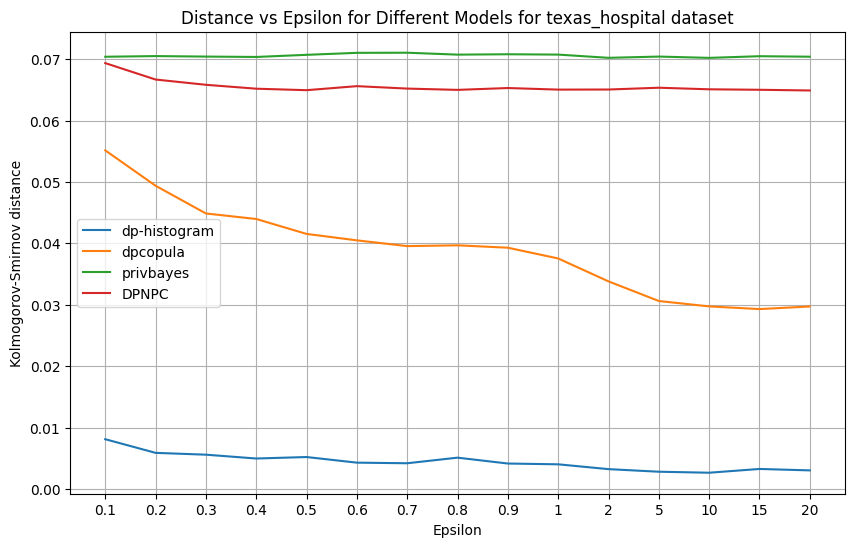}
        \caption{}
        \label{fig:fidelity_texas}
    \end{subfigure}

    \caption{Fidelity measures (a)\textit{Adult}, (b)\textit{Compas} and (c)\textit{Texas Hospital} dataset, with the KS Distance metric on the y-axis and different $\epsilon$ values along the x-axis. The closer the metric is to 0, the better the fidelity}
    \label{fig:fidelity_combined}
\end{figure*}

Regarding the privacy metric, as shown in Figure \ref{fig:fidelity_combined}, it is evident that the DP-Copula method maintains a smaller distance between the marginal distributions. It is important to note that the KS distance in this context is measured as an average of the marginal distributions, making it logical that models such as DP-Histogram and DP-Copula, which best preserve this distance, would perform well. PrivBayes and DPNPC exhibit similar behavior for the \textit{Adult} and \textit{Texas Hospital} datasets, with a notable difference in the \textit{Compas} dataset. This discrepancy may be due to the smaller number of samples in this dataset, where the PrivBayes model converges more quickly than DPNPC in terms of the number of samples required.

For this metric, we conducted a series of t-tests to compare different pairs of observations within the three datasets, adjusting the significance level for multiple comparisons using the Bonferroni correction with a threshold of $p < 0.05$. The t-tests were used to determine whether the means were significantly different. The results presented in this manuscript for the distance metric show the mean of the experiments, as no statistically significant differences were observed among the experiments.

\subsubsection{Privacy}
\begin{figure*}[h]
    \centering
    \begin{subfigure}[b]{0.45\linewidth}
        \centering
        \includegraphics[width=\linewidth]{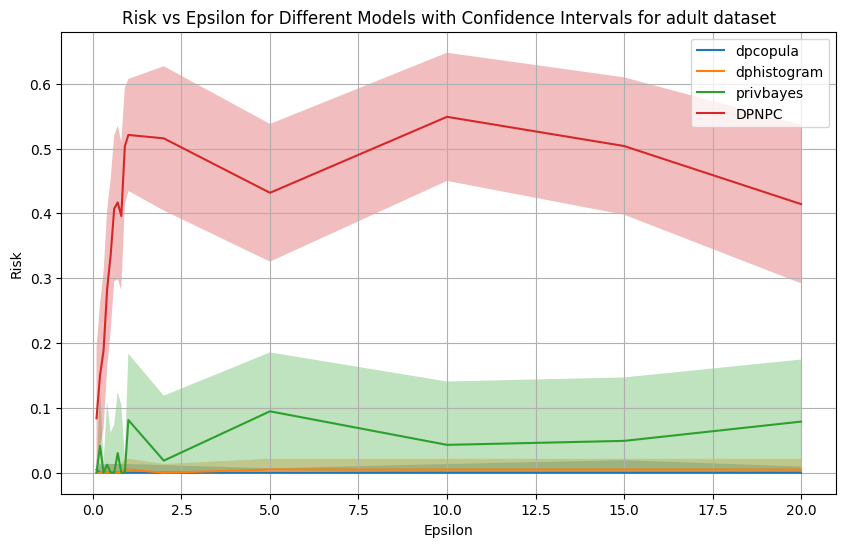}
        \caption{}
        \label{fig:risk_adult}
    \end{subfigure}
    \hfill
    \begin{subfigure}[b]{0.45\linewidth}
        \centering
        \includegraphics[width=\linewidth]{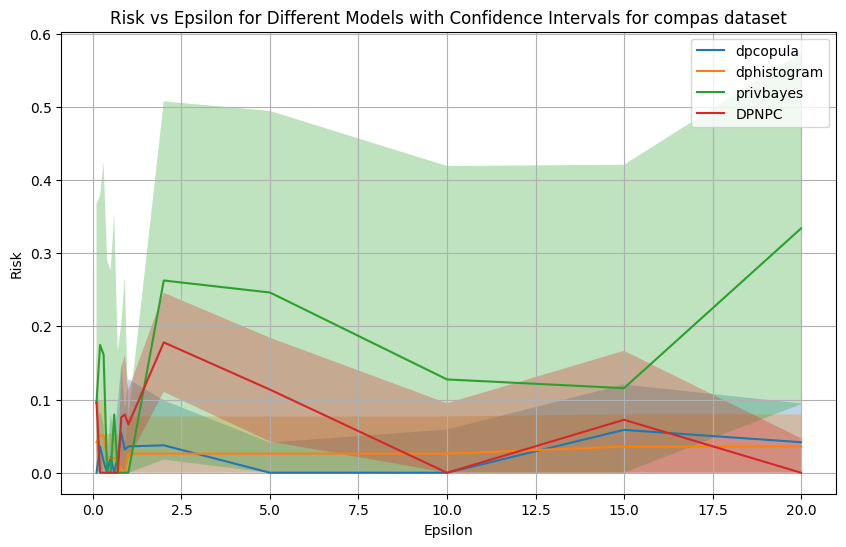}
        \caption{}
        \label{fig:risk_compas}
    \end{subfigure}
    
    \vspace{1cm}
    \begin{subfigure}[b]{0.45\linewidth}
        \centering
        \includegraphics[width=\linewidth]{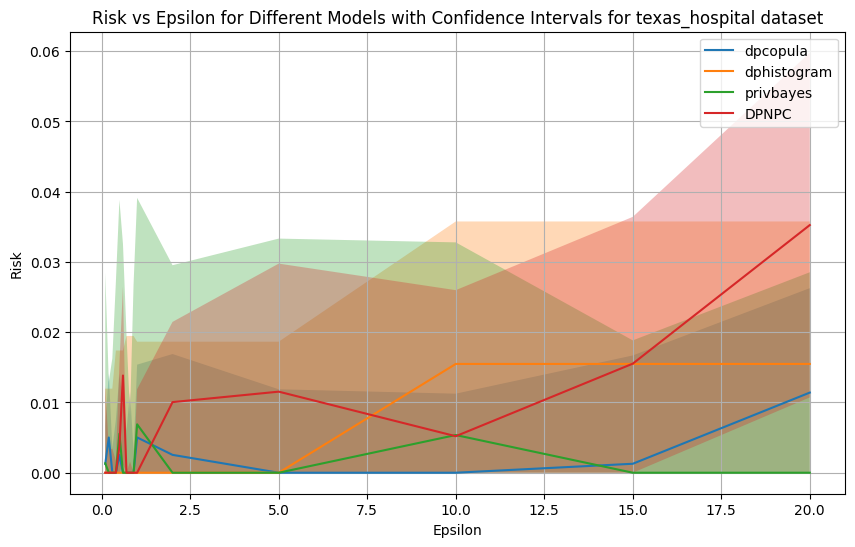}
        \caption{}
        \label{fig:risk_texas}
    \end{subfigure}

    \caption{Risk measures (a)\textit{Adult}, (b)\textit{Compas} and (c)\textit{Texas Hospital} dataset, with the risk metric on the y-axis, with a confidence interval of 95\% and different $\epsilon$ values along the x-axis. The closer the metric is to 0, the better the model respond to a MIA attack}
    \label{fig:risk_combined}
\end{figure*}

The risk metric for MIA, as shown in Figure \ref{fig:risk_combined}, demonstrates similar behavior for very small values of $\epsilon$, taking into account the confidence interval. Notably, for the \textit{Adult} dataset, the DPNPC method exhibits a high risk for $\epsilon$ values greater than 1, indicating sensitivity to privacy for very high $\epsilon$ values. For the \textit{Compas} dataset, the risk increases with models like PrivBayes, which, as noted in the previous section, better maintains the distance between synthetically generated data.

The risk for large datasets, such as \textit{Texas Hospital}, is very low, as the properties of indistinguishability are better preserved, not only due to Differential Privacy (DP) but also because of the large number of samples. For small $\epsilon$ values, the behavior of PrivBayes compared to DPNPC shows that the latter is more reliable, although this changes for larger $\epsilon$ values.

\subsection{Execution times}
\begin{figure*}[h]
    \centering
    \begin{subfigure}[b]{0.45\linewidth}
        \centering
        \includegraphics[width=0.85\linewidth]{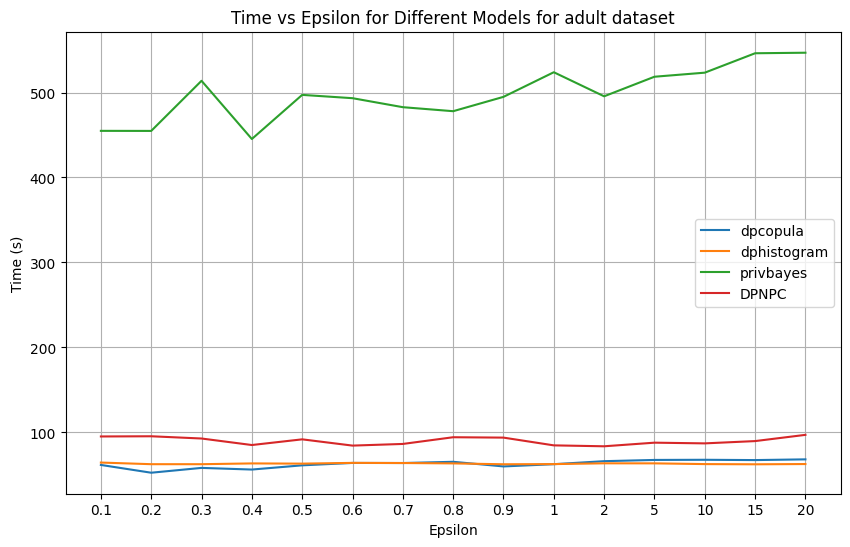}
        \caption{}
        \label{fig:time_adult}
    \end{subfigure}
    \hfill
    \begin{subfigure}[b]{0.45\linewidth}
        \centering
        \includegraphics[width=0.85\linewidth]{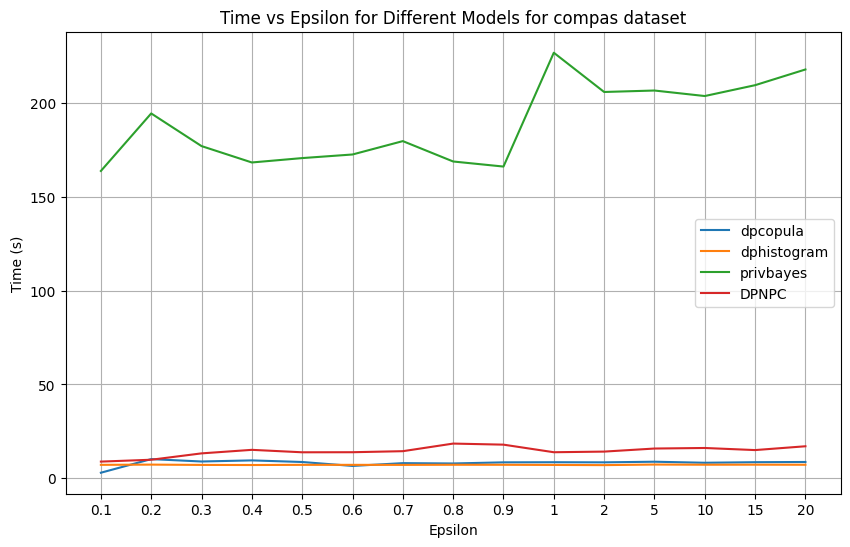}
        \caption{}
        \label{fig:time_compas}
    \end{subfigure}
    
    \vspace{1cm}
    \begin{subfigure}[b]{0.45\linewidth}
        \centering
        \includegraphics[width=0.85\linewidth]{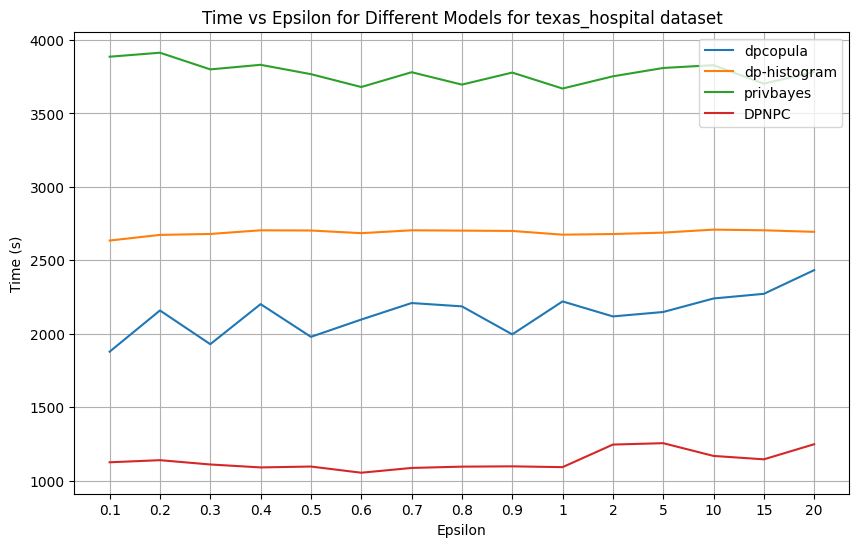}
        \caption{}
        \label{fig:time_texas}
    \end{subfigure}

    \caption{Time measures (a)\textit{Adult}, (b)\textit{Compas} and (c)\textit{Texas Hospital} dataset, with the time in seconds in the y-axis and different $\epsilon$ values along the x-axis.}
    \label{fig:time_combined}
\end{figure*}

Execution times, as illustrated in Figure \ref{fig:time_combined}, were measured by considering the duration of the pipeline execution for each dataset and $\epsilon$ value, along with the fidelity and utility metrics. It is evident that the training time for PrivBayes is significantly higher across all datasets compared to the other methods. Additionally, for large datasets such as \textit{Texas Hospital}, DPNPC demonstrates a notably shorter execution time, even when compared to models like DP-Histogram and DP-Copula.

\section{Conclusions}
This work involves the design, implementation, and comparison of a synthetic data generation algorithm with privacy guarantees. It extends the synthetic data generation model based on non-parametric copulas for mixed tabular databases by incorporating Differential Privacy through an Enhanced Fourier Perturbation method. The comparison is conducted using three public datasets and involves three synthetic data generation models: PrivBayes, DP-Copula, and DP-Histogram.
Through an experiment analyzing the resolution parameter of the method (number of \textit{bins}) in relation to the amount of noise introduced by Differential Privacy ($\epsilon$), we were able to verify the model's stability. This includes its performance in generating synthetic data with respect to distance and privacy metrics. Such analysis enables the identification of an optimal trade-off between the privacy guarantees offered by the model and the fidelity of the generated data through an appropriate combination of parameters. The utility metric demonstrates the superior performance of DPNPC in modeling the multivariate dependency structure of the data, outperforming other models. Additionally, the fidelity metric highlights the need for a significant sample size for DPNPC to achieve competitive results compared to PrivBayes.
The privacy risk measured through Membership Inference Attacks indicates that the models with the highest risk of privacy breaches, according to previous metrics, are DPNPC and PrivBayes, with few exceptions. The performance of DPNPC for $\epsilon$ values less than one is competitive in most cases, maintaining a smaller confidence interval compared to other methods. However, for very large $\epsilon$ values, other models exhibit improved performance. Finally, there is a significant difference in execution times among the methods, with PrivBayes standing out due to its substantially higher training times. In contrast, the other methods exhibit much shorter training times, even for datasets with a large number of records, with DPNPC emerging as the most efficient candidate in such scenarios.

Thus, DPNPC stands out due to its reduced training time, stable performance across variations in the \textit{bins} parameter, and effective maintenance of data utility for large datasets. Additionally, it performs well in preserving privacy guarantees for $\epsilon$ values less than 1, likely due to the efficient use of the privacy budget internally.

This study has some limitations. First, the capabilities of the predictors should be evaluated across other attributes, as the model's sensitivity is crucial when assessing utility. Additionally, relying on a single type of attack to measure privacy risk may introduce biases in evaluating how each model preserves privacy, as some models may perform better or worse depending on the type of attack used. This consideration also applies to other metrics. Furthermore, it is important to recognize the sensitivity of copula-based models to the encoding method implemented for non-continuous data. A future line of research should involve evaluating the performance of the DPNPC model with different encoding methods.





\begin{acks}
The authors used COPILOT to revise the text in the introduction section to correct any typos, grammatical errors, and awkward phrasing
\end{acks}

\bibliographystyle{ACM-Reference-Format}
\bibliography{sample-base}

\end{document}